\documentclass[final]{cvpr}

\usepackage{times}
\usepackage{epsfig}
\usepackage{graphicx}
\usepackage{amsmath}
\usepackage{amssymb}
\usepackage{booktabs}
\usepackage{xfrac}
\usepackage{multirow}
\usepackage[dvipsnames]{xcolor}

\usepackage[pagebackref=true,breaklinks=true,colorlinks,bookmarks=false]{hyperref}

\def\confYear{CVPR 2021}

\begin{document}

\title{SG-Net: Spatial Granularity Network for\\ One-Stage Video Instance Segmentation}

\author{Dongfang Liu$^{1*}$, Yiming Cui$^{2}$\thanks{indicates equal contributions.}, Wenbo Tan$^{3}$, Yingjie Chen$^{1}$\\
$^{1}$Purdue University, USA  \\
$^{2}$University of Florida, USA  \\
$^{3}$Hangzhou Dian Zi University, China\\
{\tt\small \{liu2538,victorchen\}@purdue.edu, \tt\small  cuiyiming@ufl.edu, \tt\small  174060003@hdu.edu.cn}
}

\maketitle

\begin{abstract}
Video instance segmentation (VIS) is a new and critical task in computer vision. To date, top-performing VIS methods extend the two-stage Mask R-CNN  by adding a tracking branch, leaving plenty of room for improvement. In contrast, we approach the VIS task from a new perspective and propose a one-stage spatial granularity network (SG-Net). Compared to the conventional two-stage methods, SG-Net demonstrates four advantages: 1) Our method has a one-stage compact architecture and each task head (detection, segmentation, and tracking) is crafted interdependently so they can effectively share features and enjoy the joint optimization; 2) Our mask prediction is dynamically performed on the sub-regions of each detected instance, leading to high-quality masks of fine granularity; 3) Each of our task predictions avoids using expensive proposal-based RoI features, resulting in much reduced runtime complexity per instance; 4) Our tracking head models objects’ centerness movements for tracking, which effectively enhances the tracking robustness to different object appearances. In evaluation, we present state-of-the-art comparisons on the YouTube-VIS dataset. Extensive experiments demonstrate that our compact one-stage method can achieve improved performance in both accuracy and inference speed. We hope our SG-Net could serve as a strong and flexible baseline for the VIS task. Our code will be available here\footnote{https://github.com/goodproj13/SG-Net}.
\end{abstract}
\section{Introduction}
\begin{figure}[!ht]
	\centering
	\includegraphics[width=8cm]{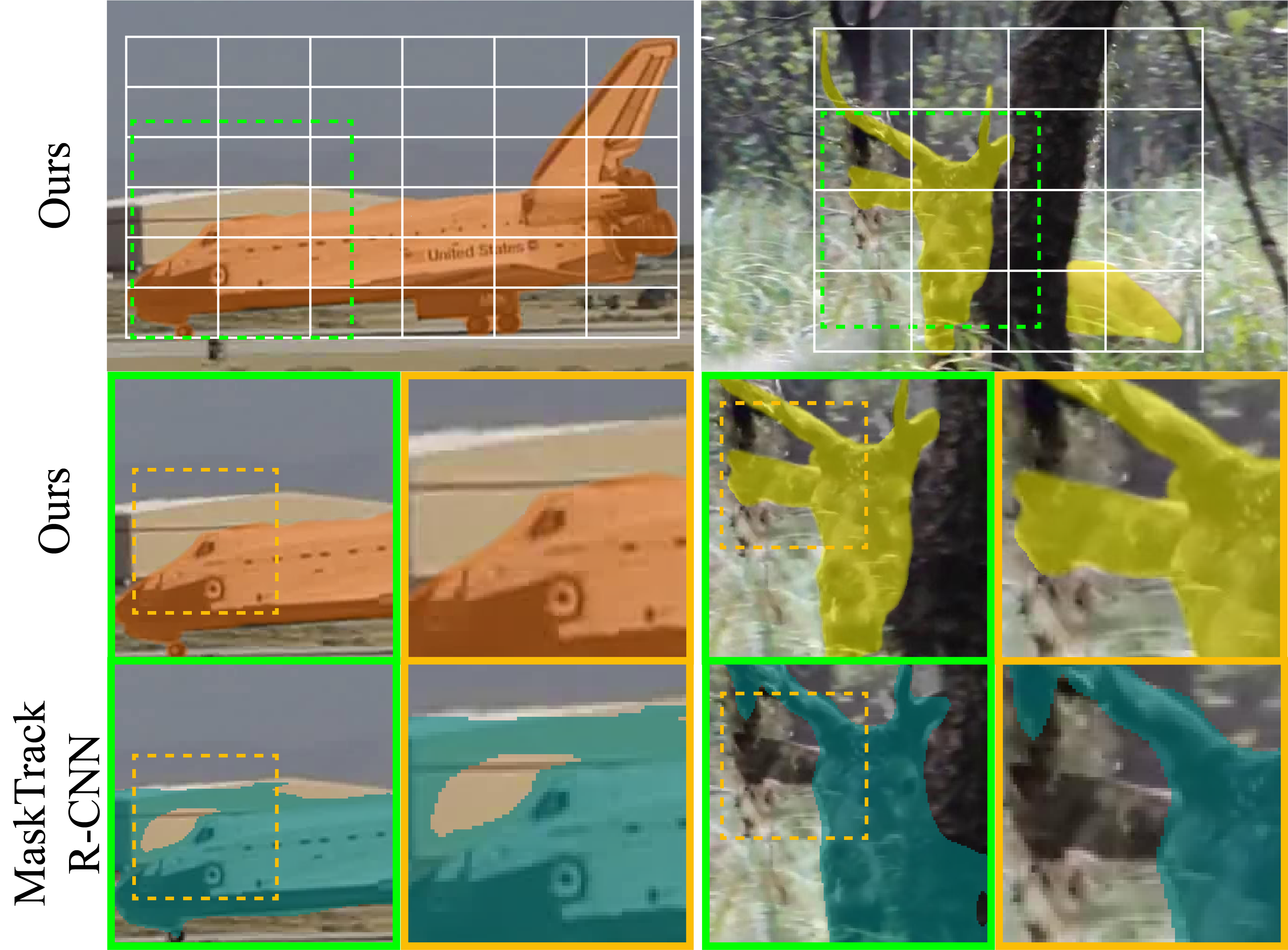}
	\caption{
	A detailed comparison with MaskTrack R-CNN. The large images in the first row are the results of our method. We further zoom in our results and compare against results from MaskTrack R-CNN. We dynamically divide a target instance into sub-regions based on its bounding box and perform instance segmentation on each sub-region to achieve spatial granularity. Both examples are cropped from the original images for better illustration.} 
	\label{first}
\end{figure}
\indent Video instance segmentation  (VIS) is a challenging vision task introduced by \cite{yang2019video}. Given a video frame, an algorithm aims to perform the task of detection, segmentation, and tracking of instances simultaneously. The VIS task holds valuable possibilities for applications which requires video-level object masks and temporal descriptions such as augmented reality, video editing, and autonomous driving.

\indent The seminal work, MaskTrack R-CNN~\cite{yang2019video} follows a two-stage paradigm as it is extended from Mask R-CNN \cite{he2017mask}. MaskTrack R-CNN first uses region proposal network (RPN) from Faster R-CNN~\cite{ren2015faster} to produce a set of candidate proposals. Then, regions-of-interests (RoI) features based on the proposals are cropped out and fed into each task head to predict bounding boxes, instance masks, and object trackings respectively. Despite a few works being proposed recently, the dominant VIS frameworks \cite{feng2019dual,kim2020video,lin2020video,liu2019spatio,mohamed2020instancemotseg} follow the two-stage paradigm.\\ 
\indent However, the two-stage paradigm may encounter a number of issues. First, it is difficult for each sub-task head (detection, segmentation, and tracking) in the two-stage approach to share features, which causes a trouble for network architecture optimization. Second, the cropped RoI features are resized into patches of a uniformed size (e.g.,  $14\times14$ or $28\times28$ in Mask R-CNN~\cite{he2017mask}), which restricts the output resolution of the instance mask. This practice could particularly compromise the prediction of large instances as they need higher resolutions to retain details of the object boundary. Third, the candidate proposals are redundant representations because their numbers are much larger than the final predictions. The mask head and tracking head have to repeatedly encode the proposal-based RoI features for the final predictions. Specifically, the mask head requires a stack of convolutions (e.g., four $3\times3$ convolutional layers in Mask R-CNN~\cite{he2017mask}) to obtain a receptive field large enough to understand sufficient image context. Therefore,  the inference runtime is largely depended on the number of detected objects that appeared on the video frame and would degrade considerably with the increase of predictions.\\
\indent Considering the above limitations, we attempt to solve the VIS task from a new perspective. We treat all three sub-tasks, detection, segmentation and tracking in VIS as interconnected problems that should be considered interdependently. Recent advances in object detection illuminate that one-stage methods such as FCOS~\cite{tian2019fcos} and CenterNet ~\cite{zhou2019objects} can outperform their two-stage counterparts in accuracy. Both methods use flexible convolutional operations, which are convenient for multi-task implementations and multi-network optimization (i.e., instance segmentation and tracking). Their successes enable the possibility to implement a one-stage approach for the VIS task.\\
\indent For the instance segmentation task, we recognize two important predecessors, BlendMask \cite{chen2020blendmask} and CondInst \cite{tian2020conditional}, which  are both built on the FCOS framework. Compared to the Mask R-CNN-based approaches, they only use fully convolutional network (FCN) architectures, which help them to avoid the RoI operations (i.e., RoIpool, RoIAlign, cropping, and resizing). This improvement significantly preserves the feature map resolution and retains details of the mask boundary. Moreover, both methods implement a light-weight mask head, which makes them very robust in real-time video tasks. However, both methods segment objects on the instance level and ignore the potential to achieve more granularity on objects.\\ 
\indent 
In this work, our primary focus is to investigate ways to circumvent the heavy proposal-based two-stage approach and find a flexible solution for the VIS task.
Inspired by~\cite{chen2020blendmask,tian2020conditional}, 
we generalize the mask prediction by
modeling spatial scales of objects with semantic information for lower-level granularity. We also propose an easy track strategy, which directly uses the object centerness from detection to delineate the video temporal coherence.
Concretely, our work delivers the following contributions:
\begin{itemize}
    \item We attempt to solve the VIS task from a new perspective. To this end, we devise a compact one-stage method called SG-Net. Our method dynamically divides instance into sub-regions and performs segmentation on each region for \textbf{S}patial \textbf{G}ranularity, thus the name of our network, SG-Net. Compared to MaskTrack R-CNN~\cite{yang2019video}, 
    our method achieves more appealing segmentation behavior as it can enrich object details and produce masks with more accurate edges (As shown in Figure~\ref{first}). 
    \item Our method is proposal-free and efficient.  Removing proposals allows us to assign heavier duties to the mask prediction module with an affordable computation overhead.  Particularly, our inference time does not increase with the growing number of predictions as the two-stage methods do.
    \item Our entire architecture consists of only convolutional operations, which is tied to the state of the art one-stage object detector, FCOS~\cite{tian2019fcos}.  We organically craft each task head interdependently so they can effectively share features and enjoy joint optimization.
    \item Our tracking head models the movement of object centerness for tracking, which is simple and effective. Compared to MaskTrack R-CNN~\cite{yang2019video}, our tracking head is more robust to different object shapes and sizes as well as appearance changes. 
\end{itemize}
\begin{figure*}[!ht]
	\centering
	\includegraphics[width=\textwidth]{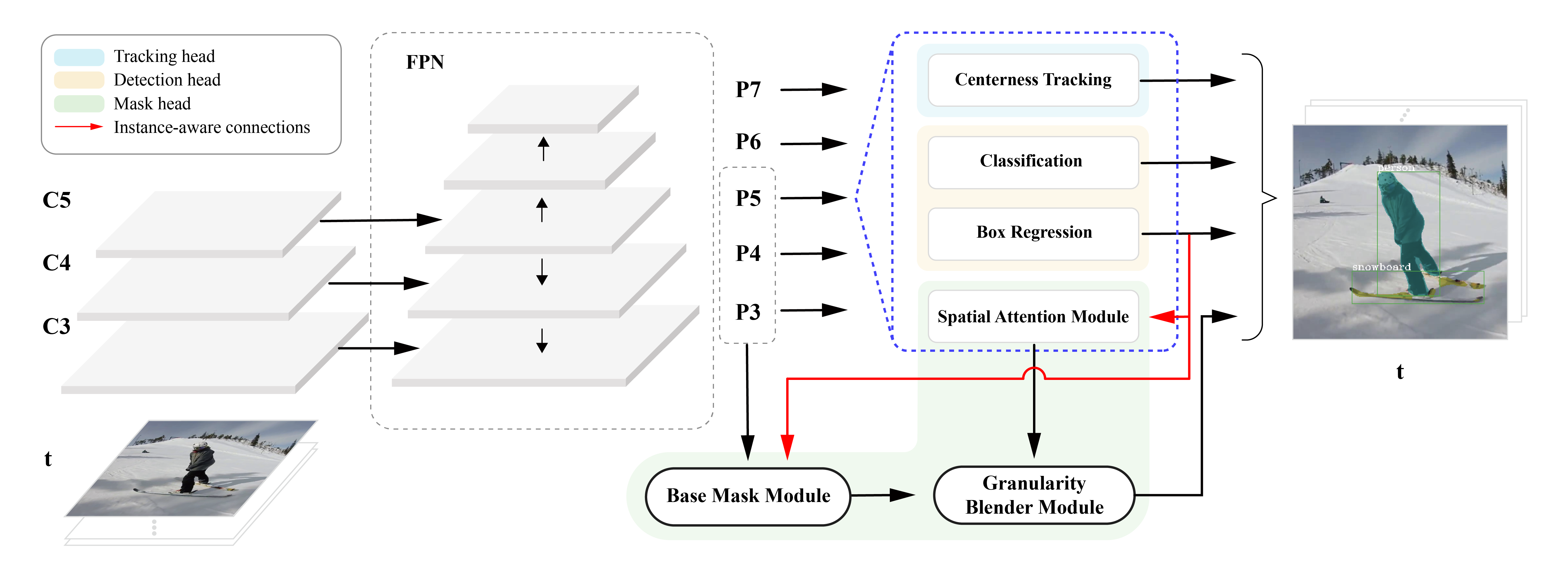}
	\caption{Illustration of SG-Net framework. Our detection head includes the classification branch, the box regression branch, the centerness branch (not shown for simplicity) as in FCOS~\cite{tian2019fcos}. Our tracking head directly uses the centerness outputs from FCOS to model object movements for tracking. We use the instance-aware connections to propagate the bounding box information and guide the instance segmentation task. Therefore, our mask head can efficiently encode on the pixel of the instance, instead of repeatedly performing predictions on proposal-based features as in two-stage methods. Specifically, the spatial attention module divides the target instance into equally-sized sub-regions based on its size and shape and predict an attention score for each sub-region. The base mask module produces a set of base masks which is equal to the number of the sub-regions of the same instance. The granularity blender  combines the attention scores and the base masks to predict the final mask. Note that the functional modules inside of the blue dashed box are repeatedly applied to $\{P3,...,P7\}$.}
	\label{Frame}
\end{figure*}
\section{Related Work}
Although VIS is a new task in computer vision, its sub-tasks (a.k.a., detection, instance segmentation, and tacking) have been well studied in the literature. This section summarizes the related work in these tasks.
\subsection{Object Detection}
One-stage detectors can achieve a faster inference speed by avoiding proposal generation \cite{liu2016ssd,redmon2016you}. However, one-stage approaches are generally inferior in accuracy compared to their two-stage counterparts. The latest one-stage object detection methods (i.e, FCOS~\cite{tian2019fcos} and CenterNet~\cite{zhou2019objects}) prove the effectiveness of removing box anchors to simplify the detection pipeline. Without any restriction of the pre-defined anchors, target objects are matched freely to the prediction features of their suitable receptive fields. Compared to its anchor-based counterparts~\cite{lin2017focal,liu2016ssd,redmon2016you}, FCOS~\cite{tian2019fcos} and CenterNet~\cite{zhou2019objects} not only improve the detection accuracy, but also remain a fast inference. The flexibility of one-stage detection framework benefits the instance segmentation methods \cite{chen2020blendmask,tian2020conditional} as it removes the computation overhead caused by proposal or anchor operations. Building on the lessons learned from the concurrent approaches \cite{chen2020blendmask,tian2020conditional}, our approach is based on the one-stage FCOS framework.
\subsection{Instance Segmentation}
To date, the dominant instance segmentation methods still follow the two-stage paradigm based on Mask R-CNN\cite{he2017mask}. These methods \cite{chen2019hybrid,huang2019mask,liu2018path} rely heavily on RoI features and operations, which first produce a set of candidate proposals and then predict the foreground masks on each of the RoIs.  Although there are several attempts been made for using alternative  approaches \cite{dai2016instance,fathi2017semantic,neven2019instance,newell2017associative,novotny2018semi}, Mask R-CNN still outperforms these methods in accuracy.\\
\indent More recently, YOLACT~\cite{bolya2019yolact}, SipMask~\cite{cao2020sipmask}, and BlendMask~\cite{chen2020blendmask} reformulate the instance segmentation  pipeline based on Mask R-CNN. YOLACT~\cite{bolya2019yolact} and SipMask~\cite{cao2020sipmask} generate a number of prototype masks for each instance and combine them with per-pixel predictions using a single coefficient for the entire instance
(YOLACT) and $2\times2$ coefficients for each sub-region of the instance (SipMask), respectively. However, these overly simplified assembling designs may not achieve the optimal representation of the object features. Similarly, BlendMask~\cite{chen2020blendmask} first predicts base masks and corresponding 2D attention maps for each instance, and then blends them together for the final prediction. Since BlendMask performs segmentation directly at the instance level, its representation of object details is not sufficient. Inspired by the aforementioned methods ~\cite{cao2020sipmask,chen2020blendmask,tian2020conditional}, our approach models objects’ spatial scales with semantic information. Based on objects' sizes and shapes, we dynamically divide them into different sub-regions for segmentation to achieve improved mask granularity.
\subsection{Multiple Object Tracking}
Recent multi-object tracking mainly follows the tracking-by-detection paradigm \cite{fang2018recurrent,leal2017tracking,schulter2017deep,xu2019spatial,zhu2018online}, which first detects objects and then predicts the temporal association of all detections. Some methods use motion priors, such as Kalman Filter~\cite{feng2019multi,yu2016poi} or optical flow~\cite{xiao2018simple} to associate the detected bounding boxes. Others learn tracking representations by exploiting the appearance similarity \cite{sadeghian2017tracking,xu2019spatial,son2017multi} or re-identify vanished objects \cite{leal2016learning,tang2017multiple,tang2019cityflow}. Our tracking also follows the tracking-by-detection paradigm. Instead of using the object bounding box or mask as the  tracking clue, we track object movement by its center, which is the output from the centerness branch of FCOS~\cite{tian2019fcos}. Therefore, we can optimize detection and tracking jointly and achieve an improved tracking behavior. 
\subsection{Video Instance Segmentation}
Majority of VIS methods~\cite{bertasius2020classifying,dong2019temporal,feng2019dual,kim2020video,lin2020video,liu2019spatio,mohamed2020instancemotseg} follow the  two-stage paradigm based on Mask R-CNN. Although the  two-stage paradigm is intuitively simple, it has several performance bottlenecks: 1) feature sharing cross task heads is insufficient so joint optimization of network architecture is difficult; 2) the quality of the instance mask is restricted by the RoI operation (e.g., resizing or pooling); 3) the RoI feature representation is redundant so the  inference speed is largely effected by the number of  candidate proposals. Compared to the two-stage methods, our SG-Net is fast and one-stage. Our mask head is fully-convolutional (no RoI operations) and proposal-free (no redundant RoI features), so we can produce high-quality instance masks while achieving a fast inference. Each of our task heads is crafted interdependently and can be jointly optimized. 
\section{Our SG-Net}
\subsection{Overall Architecture}
Built on the FCOS~\cite{tian2019fcos}, our SG-Net includes a feature extraction backbone, a detection head, a mask head, and a tracking head.  Our overall architecture is shown in Figure~\ref{Frame}. Our backbone network employs ResNet~\cite{he2016deep} with FPNs~\cite{lin2017feature}. We exploit the feature maps from \{$P3, P4, P5, P6, P7$\} of FPNs and apply different functional modules to fulfill different sub-tasks. Our detection  adopts the original FCOS~\cite{tian2019fcos}, which consists of the classification, box regression, centerness branch. At time $t$, we denote the detection of the $i^{th}$ object as $D^{i}_t=(c^i_t, o^i_t, b^i_t),$ where $c^i_t\in \{0,...,C-1\}$ is the class of the detected object, $o^i_t\in\mathbb{R}^2$ is the centerness location of the detected object, and $b^i_t\in\mathbb{R}^4$ is the object bounding box. Refer to~\cite{tian2019fcos} for more details. Since our mask and tracking head are the key components of SG-Net, we elaborate the two techniques in the following sections.
\subsection{Mask Head} \label{maskHeadSection}
\begin{figure}[!ht]
	\centering
	\includegraphics[width=8cm]{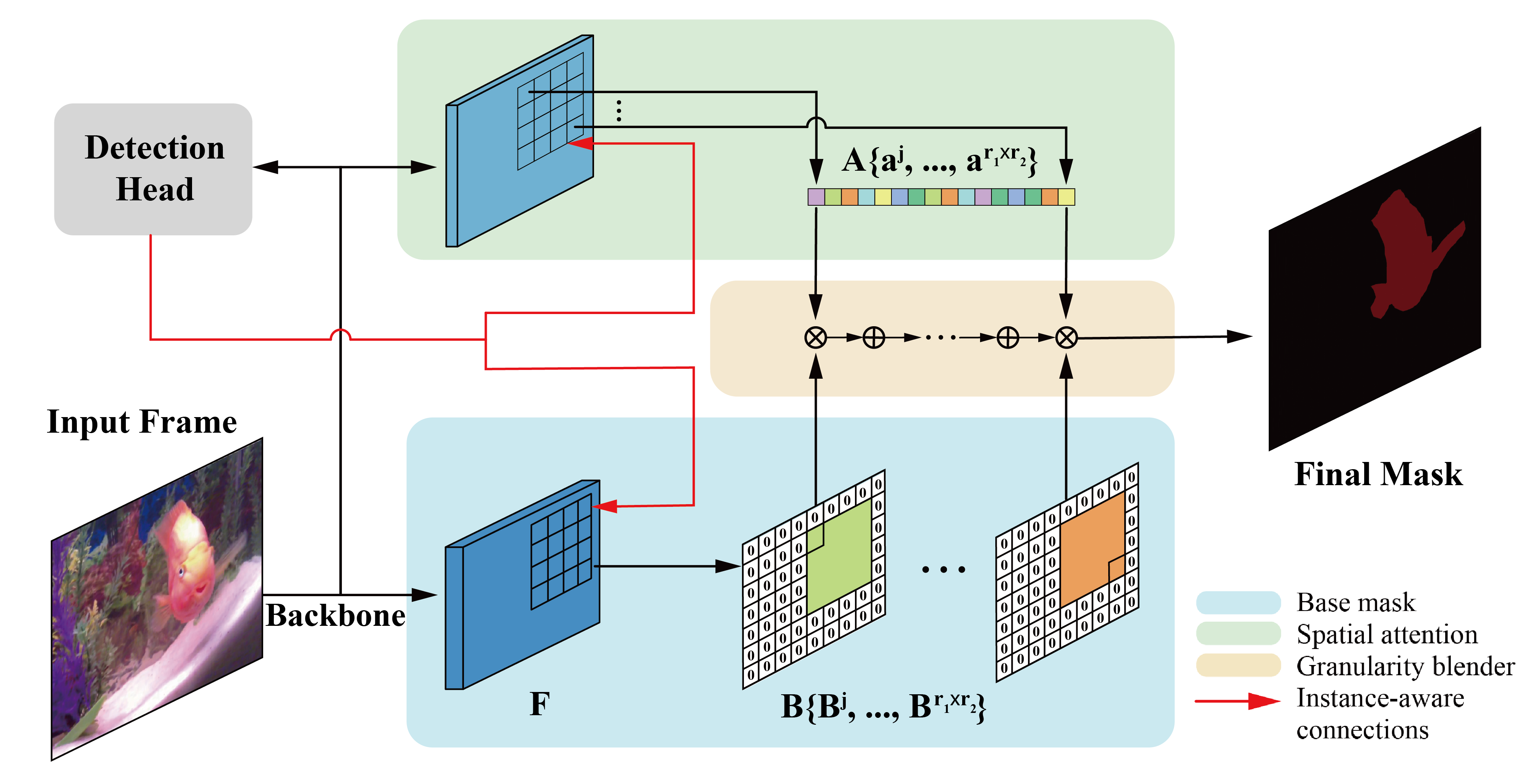}
	\caption{The working pipeline of our mask head.}
	\label{blender}
\end{figure}

Our mask head consists of: 1) a spatial attention module; 2) a base mask module; 3) a granularity blender. Each module works collaboratively to predict the final mask.\\
\indent \textbf{Spatial attention
module:} 
Our spatial attention module divides the object bounding box into multiple sub-regions and predicts attention scores on each sub-region (the green area in Figure~\ref{blender}). We use the instance-aware connection to propagate the bounding box information \{$b^1,..,b^n$\} from detection results to the spatial attention module. Based on objects' shapes and sizes, we dynamically divide the object bounding box into $r_1\times r_2$ sub-regions:
\begin{equation}
r_1,~r_2=\min(6,\lceil \frac{w}{50} \rceil),~\min(6,\lceil \frac{h}{50} \rceil)
\label{sub}
\end{equation}
where $w$ and $h$ are the width and height of the bounding box respectively and the number 50 is in a pixel unit. Instead of pooling features into 3D parameter-heavy attention maps in  BlendMask~\cite{tian2020conditional}, we predict a set of 1D attention scores (similar to coefficients in  YOLACT~\cite{bolya2019yolact}) for every sub-region by appending two $3\times3$ convolution layers to each of the $P3-P7$ levels. For a detected object, we define its attention scores as $\mathbb{A}=\{a^j\in \mathbb{R}|j=1,...,r_1\times r_2\}$ where $a^j$ is the attention score for the $j^{th}$ sub-region. 
\\
\indent \textbf{Base mask module:} The base mask generation pipeline has two major steps: 1) extract and prepare FPNs features; 2) generate the base masks for each target instance. In the first step, we first extract the $P3-P5$ features from FPNs. $P4-P5$ features are upsampled to the same  resolution of $P3$ using bilinear interpolation. Afterward, the three-level features ($P3-P5$) are concatenated together and fed into two $3\times3$ convolutions with 128 channels. In order to reduce the computational parameters, we append another convolution layer to downsize the feature channels from 128 to 36. The outputs from the previous step are upsampled $\times4$ using bilinear interpolation to produce high-resolution feature maps $F$. Our later evaluations show that the feature upsampling is beneficial to the final mask prediction.\\
\indent In the second step, we use the instance-aware connection to guide the base mask production (the blue area in Figure ~\ref{blender}). Informed by the bounding box locations \{$b^1,..,b^n$\}, we can perform prediction accurately within the region of the detected object. To produce the base masks for a specific instance, we first assign the local features outside of its bounding box to be zero and obtain new target-oriented feature maps, where we apply a $1\times1$ convolution to dynamically produce our base masks. The number of the base masks for an instance is restricted by the  volume of  its sub-regions ($r_1\times r_2$) from the spatial attention module. For a detected object, we define its base masks as $\mathbb{B}=\{B^j\in \mathbb{R}^{\frac{H}{2}\times \frac{W}{2}}|j=1,...,r_1\times r_2\}$ where $B^j$ is the the $j^{th}$ base mask and $\frac{H}{2}\times \frac{W}{2}$ is the base mask resolutions ($H\times W$ is the input size). The spatial attention module and base mask module progress in parallel. Therefore, each base mask $B^j$ has its corresponding attention score $a^j$.\\
\indent \textbf{Granularity blender module:}
Our granularity blender takes the attention scores and the base masks as inputs to produce the final mask (the yellow area in Figure~\ref{blender}). For the $i^{th}$ detected object on the frame, we perform a summation of an element-wise multiplication between its attention scores $\mathbb{A}$ and base masks $\mathbb{B}$ to predict its instance mask $M^i$:
\begin{equation}
    M^i = \sum_{j=1}^{r_1\times r_2}sigmoid(a^j\times B^j),
    \label{Equation 1}
\end{equation}
where $sigmoid()$ is to binarize the summation. Accordingly, we can obtain the instance masks $M$ for the whole frame as the follows:   
\begin{equation}
    M = \sum_{i=1}^{n}M^i, 
    \label{Equation 1}
\end{equation}
where $n$ is the total number of predictions on the frame.\\
\indent At the first glance, our mask head may incur a large number of network parameters, when processing video frames with a high volume of objects. However, we circumvent inefficient proposal generations and  RoI operations in two-stage methods, resulting in much-reduced runtime complexity per instance and improved segmentation behavior.
\subsection{Tracking Head}
\begin{figure}[!ht]
	\centering
	\includegraphics[width=8cm]{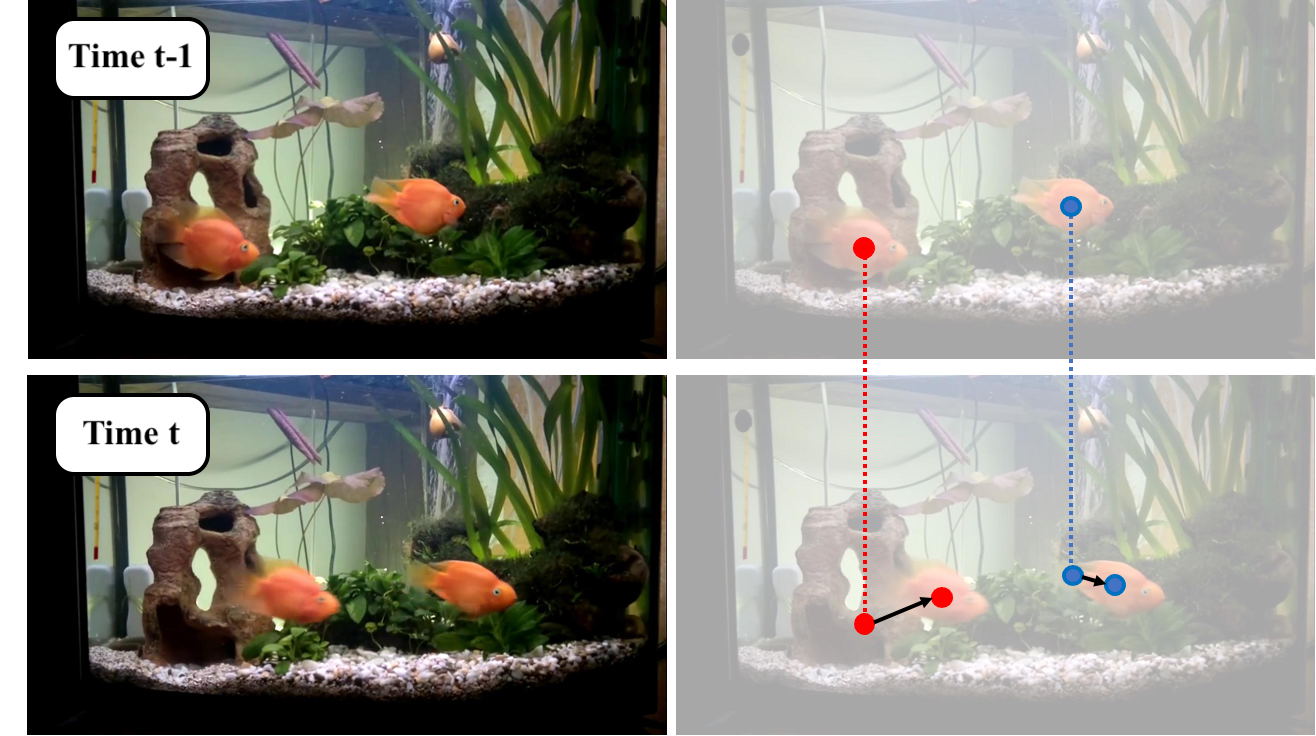}
	\caption{The centerness tracking. We use object centers to model object movements. By calculating object centerness movement across frames, we can depict the video temporal coherence.}
	\label{centertrack}
\end{figure}

Our tracking head uses centerness to represent each instance. Since FCOS already has a centerness branch, we can conveniently craft our tracking head into the FCOS framework. After FCOS's centerness branch, we append one $1\times1$ convolutional layer to process the centerness features. The work similar to us is CenterTrack~\cite{zhou2020tracking} using box centers. Our centerness tracking is more robust to object appearance changes, while box centers can reside at the same position in multiple bounding boxes, causing confusion in tracking.\\
\indent At time $t$, 
tracking for one object $t_t=(o_t, id_t)$ is described by its centerness location $o_t \in \mathbb{R}^2$ from the detection result and a unique identity $id_t \in [0,  \tilde{N}]$, where $\tilde{N}$ is the total number of distinct instances so far. A newly detected object can only be assigned to one of the $\tilde{N}$ identities if it is one of the previous instances or a new $id$ if it is a new unique instance. To achieve this goal, 
our tracking head predicts a 2D movement $D \in \mathbb{R}^{\tilde{H}\times \tilde{W} \times 2}$ to associate centerness through time. For the $i^{th}$ detection at time $t$, $D^i_t$ describes the centerness movements $o^i_t - o^i_{t-1}$ between the current frame $t$ and the previous frame $t-1$.
Therefore, we can model the movement regression objective by:
\begin{equation}
\mathcal{L}_{track} = \frac{1}{N}\sum_{t \in T}\sum_{i \in n}\bigg|D^i_t-(o^i_t-o^i_{t-1})\bigg|,
\label{trackloss}
\end{equation}
where $n$ is the number of detections on a single frame at time $t$, and $N$ is all detections within the video time span $T$. In training, $T$ is the number of video sequences. Figure~\ref{centertrack} shows an example of this movement prediction.\\
\indent With the movement prediction, we can use a simple greedy matching to link objects across frame. For a detection $i$ at position $o^i$ of time $t$, we greedily associate its $id$ with the closest prior detection at position $o^i_{t}-D^i_t$. If no matched candidate within a radius $r$ is found, we spawn a new tracklet. We define $r$ as the mean of the bounding box width and height of the tracking object. Our simple tracking technique is empirically effective. 
\\
\subsection{Training Objective}
Our method follows the top-down paradigm. We use the bounding box locations from detection to guide the spatial attention  module and base  mask  module for instance segmentation. Therefore, we argue that the improved bounding box regression can benefit the instance segmentation task. Bear this in mind, we modify the GIoU loss ($\mathcal{L}_{GIoU}=1-GIoU$) in~\cite{rezatofighi2019generalized} to accommodate our own system design. We endeavor to facilitate the hard sample learning when dealing with small GIoU. To achieve this goal, we use a logarithmic function to increase the bounding box losses:
\begin{equation}
\mathcal{L}_{box} =-ln\frac{1+GIoU}{2}
\label{box}
\end{equation}
\indent Our overall loss is the sum of individual loss from the detection head, the mask head, and the tracking head:
\begin{equation}
\mathcal{L}_{all} = \mathcal{L}_{det}+\mathcal{L}_{mask}+\mathcal{L}_{track}
\end{equation}
where $\mathcal{L}_{mask}$ is the mask  loss as in \cite{he2017mask} and $\mathcal{L}_{track}$ is the movement loss from Eq. \ref{trackloss}. $\mathcal{L}_{det}$ include three terms $\mathcal{L}_{cls}$, $\mathcal{L}_{cent}$,  and $\mathcal{L}_{box}$. $\mathcal{L}_{cls}$ and $\mathcal{L}_{cent}$ are the focal loss and the centerness loss in  \cite{lin2017focal} respectively.  $\mathcal{L}_{box}$ is the modified GIoU loss from Eq. \ref{box}.
\section{Experiments}
We assess SG-Net on YouTube-VIS dataset~\cite{yang2019video}. Following \cite{yang2019video}, we perform training on the 3,471 training videos and testing on the 507 validation videos with the same input size $640\times360$. Our main results and ablation experiments are all reported on the validation set. 
\begin{table*}[!ht]
    
    \centering
    \begin{tabular}{c|c|c|c|c|c|c|c|c}
    \toprule
    \multicolumn{2}{c|}{Method} &Backbone&Time (ms)& AP & AP@0.5 & AP@0.75 & AR@1 & AR@10\\
    \midrule
    \multirow{8}{*}{\rotatebox{90}{Two Stage}}& OSMN MaskProp~\cite{yang2018efficient}&ResNet-50&-& 23.4 & 36.5 & 25.7 & 28.9 & 31.1\\
    & IoUTracker+~\cite{yang2019video}&ResNet-50&-& 23.6 & 39.2 & 25.5 & 26.2 & 30.9\\
    & FEELVOS~\cite{voigtlaender2019feelvos}&ResNet-50&-& 26.9 & 42.0 & 29.7 & 29.9 & 33.4\\
    & OSMN~\cite{yang2018efficient}&ResNet-50& -&  27.5 & 45.1 & 29.1 & 28.6 & 31.1\\
    & DeepSORT~&ResNet-50& -&  26.1 & 42.9 & 26.1 & 27.8 & 31.3\\
    & SeqTracker~\cite{yang2019video}&ResNet-50& -& 27.5 & 45.7 & 28.7 & 29.7 & 32.5\\
    \cline{2-9}
   \rule{0pt}{10pt} & MaskTrack R-CNN~\cite{yang2019video} &ResNet-50&98.8& 30.3 & 51.1 & 32.6 & 31.0 & 35.5 \\
  & MaskTrack R-CNN~\cite{yang2019video} &ResNet-101& 142.8& 31.8 & 53.0 & 33.6 & 33.2 & 37.6 \\
        \midrule
     \multirow{6}{*}{\rotatebox{90}{One Stage}} &STEm-Seg~\cite{athar2020stem} &ResNet-50& 83.3& 30.6 & 50.7 & 33.5 & 31.6&37.1\\
     & SipMask~\cite{cao2020sipmask} &ResNet-50& 35.7&  33.7 & 54.1 & 35.8 & 35.4 & 40.1\\
     & \textbf{Ours} &ResNet-50& \textbf{43.5} &  \textbf{34.8} & \textbf{56.1} & \textbf{36.8} & \textbf{35.8} & \textbf{40.8}\\
    \cline{2-9}
    \rule{0pt}{10pt} & STEm-Seg~\cite{athar2020stem} &ResNet-101& 125.0 & 34.6 & 55.8 & 37.9 & 34.4&41.6\\
     & SipMask~\cite{cao2020sipmask} &ResNet-101& 41.7&  35.8 & 56.0 & 39.0 & 35.4 & 42.4\\
     & \textbf{Ours} &ResNet-101&\textbf{50.6}&\textbf{36.3}&\textbf{57.1}&\textbf{39.6}&\textbf{35.9}&\textbf{43.0}\\
    \bottomrule
    \end{tabular}
    
    \label{tab:my_label}
    \caption{Quantitative results on YouTube-VIS validation set. Metrics for MaskTrack R-CNN, STEm-Seg, and SipMask are obtained using their official code and trained model. Other results are retrieved from~\cite{yang2019video}. AP and AR are the Average Precision and Recall respectively. 
    }
\end{table*}
\subsection{Implementation Details}
ImageNet pretrained ResNet~\cite{he2016deep} (the 50 and 101 version) with FPN is used as our backbone network. For the newly added layers, we initialize them as in~\cite{chen2020blendmask}. Our models are trained with SGD with momentum 0.9 and an initial
the learning rate of 0.005 which decays exponentially.
our models are trained with one TITAN RTX GPU, while all tests are conducted on a single 1080Ti GPU.
\subsection{Main Results}
\textbf{Quantitative results:} We compare our SG-Net against previous state-of-the-art methods on the validation set of YouTube-VIS. Table \ref{tab:my_label} presents the comparison results. Our
methods achieve competitive results under all evaluation metrics. In head-to-head comparisons,
our SG-Net outperforms all two-stage methods. Specifically, our margins over the state-of-the-art MaskTrack R-CNN~\cite{yang2019video} are 8.2\% and 7.9\% in AP using ResNet-50 and -101 respectively. We also compare SG-Net against the emerging one-stage methods, STEm-Seg~\cite{athar2020stem} and SipMask~\cite{cao2020sipmask}. Our SG-Net consistently surpasses STEm-Seg by a significant margin (34.8\% vs. 30.6\% using ResNet-50 and 36.3\% vs. 34.6\% using ResNet-101). Moreover, we observe steady improvements from SipMask~\cite{cao2020sipmask} (34.8\% vs. 33.7\% using ResNet-50 and 36.3\% vs. 35.8\% using ResNet-101).\\ 
\indent SG-Net is also efficient. Compared to the two-stage methods, our simpler method, without any bells and whistles, achieves both higher accuracy and faster inference speed. Our models are about $\times$2 and $\times$3 faster than MaskTrack R-CNN~\cite{yang2019video} with ResNet-50 and ResNet-101 respectively. Measured on a 1080Ti GPU, the best SG-Net (ResNet-50) typically runs at 0.043s/image, with nearly negligible impact of scene  complexity. On the contrary, the two-stage
MaskTrack R-CNN has a more expensive mask and tracking computation and its inference time closely relies on the number of predictions. Furthermore, our method achieves a faster speed than  STEm-Seg~\cite{athar2020stem}~(43.5 ms vs. 83.3 ms with ResNet-50 and 50.6 ms vs. 125 ms with ResNet-101), as it has three parameter-heavy decoders for mask generation, object tracking, and class prediction respectively. Compared to SipMask~\cite{cao2020sipmask}, our methods achieve better accuracy with on par execution time. For fair comparisons, all the inference times here are measured by the same local hardware with the official codes.
\begin{figure*}[!ht]
	\centering
	\scalebox{0.96}{\includegraphics[width=16cm]{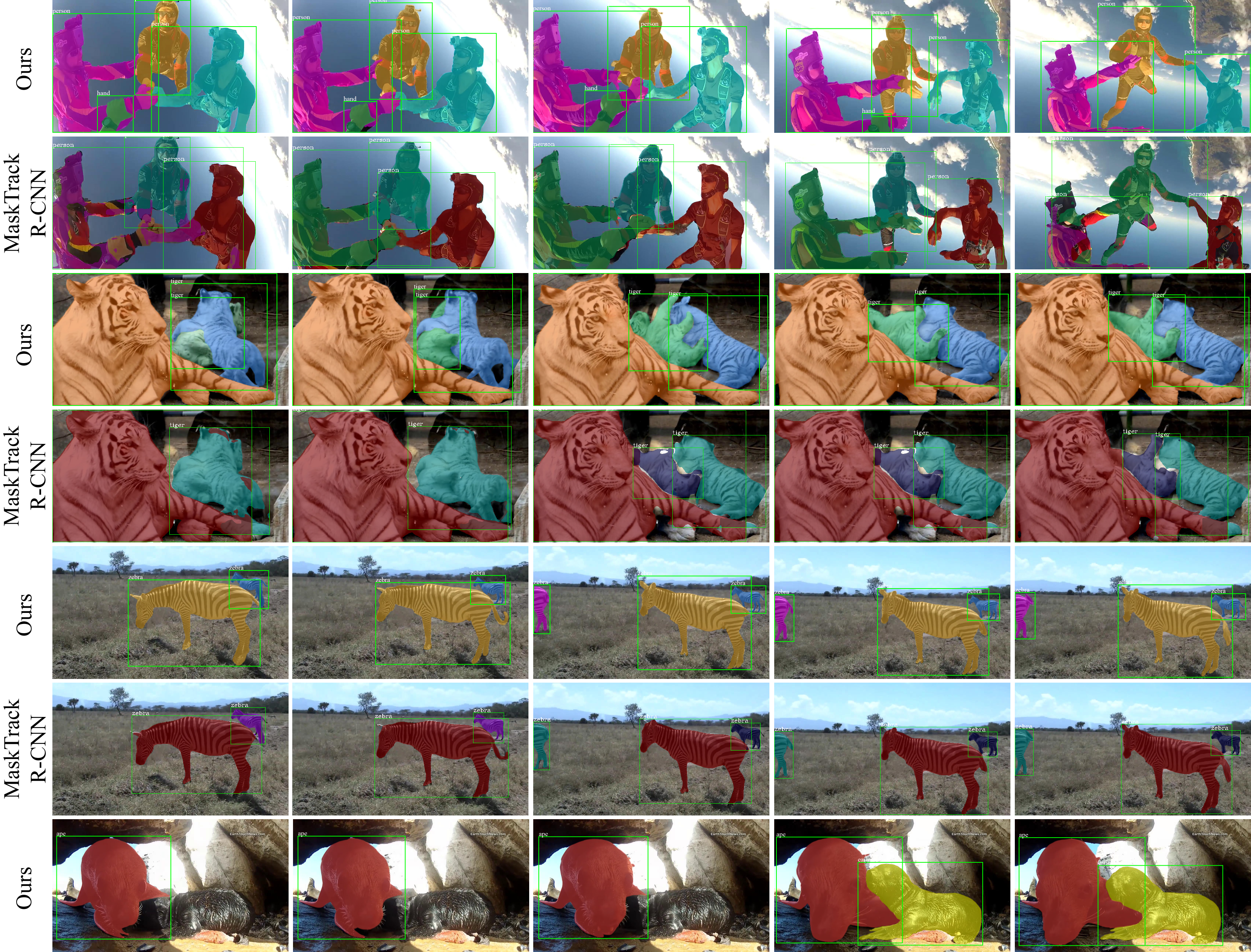}
	}
	\caption{Selected results of MaskTrack R-CNN and SG-Net. Both methods use ResNet-101 with FPN. Each row has five frames from a video sequence.  Objects with the same predicated identity have the same mask color. Best viewed in the digital format with zoom.}
	\label{track}
\end{figure*}\\
\indent \textbf{Qualitative results:}
We compare our qualitative results with those from MaskTrack R-CNN~\cite{yang2019video}. To demonstrate our advantages, we select some challenging samples where MaskTrack R-CNN struggles to deal with (as shown in Figure \ref{track}). For instance, in the skydiver examples, MaskTrack R-CNN~\cite{yang2019video} intends to assign parts to the wrong instance and leave large
false-negative regions; in the tiger examples, MaskTrack R-CNN~\cite{yang2019video} has difficulties to discriminate instances of the same class close to each other; finally, in the zebra examples, occlusions could also compromise the tracking prediction of MaskTrack R-CNN~\cite{yang2019video}.
Our method demonstrates improved segmenting and tracking behaviors on these difficult cases because: 1) Our base masks  have higher resolution (stride-2 features). Thus details are better retained. 2) Our attention module captures instance details, which guide the base masks to output results of fine granularity. 3) Our tracking head uses centerness to track objects, which is flexible  and robust to different object appearances.\\
\indent The last row in Figure \ref{track} demonstrates examples of our weak predictions. Our method fails to recognize the right earless seal for the first three frames and constantly classify the left seal as an ape. We argue that one critical reason, which causes the erroneous prediction, is the data bias on YouTube-VIS, as the “earless seal” class is underrepresented in the training set. More qualitative examples can be found here \url{https://youtu.be/zft0T3YUgpM}.
\subsection{Ablation Study}
We perform an ablation study to investigate the effectiveness of the different contributing components and configurable parameters in our method. The following results are based on ResNet101-FPN~\cite{lin2017feature}.
\begin{table}[!ht]

    \centering
    \begin{tabular}{c|c|c|c|c|c}
    \toprule
    Method&a&b&c&d&\textbf{e} \\
    \midrule
  Baseline& $\surd$ & $\surd$ & $\surd$ &$\surd$ & $\surd$  \\
  BMM& & $\surd$ &  &$\surd$ & $\surd$ \\
  SAM&  &  & $\surd$ & $\surd$ & $\surd$ \\
  GL&  & & &&$\surd$ \\
      \midrule  
    AP&  31.1 &  34.3& 35.2 & 36.1&\textbf{36.3} \\
  Time (ms)&  33.3&45.7& 51.9&50.6&\textbf{50.6}\\
    \bottomrule
    \end{tabular}
    \caption{Impact of integrating different functional  modules into the baseline to the accuracy. BMM, SAM, GL stand for base mask module, spatial attention module, and GIoU loss respectively.}
    \label{tab:cc}
\end{table}\\
\indent \textbf{Contributing components:} We first show the contributions of progressively integrating different components (Table \ref{tab:cc}): the base mask module (BMM), the spatial attention module (SAM), and the GIoU loss (GL), to the baseline. Our baseline, which is similar to YOLACT~\cite{bolya2019yolact} with our tracking head, only uses a single set of attention scores and obtain the base masks from $P3$ level of FPN. The baseline achieves an AP of 31.1\% with a speed of 33.3 ms. Each component (BMM, SAM, and GL) contributes towards improving the overall performance in accuracy. Method b adds BMM with multi-level FPN features ($P3-P5$) to the baseline and uses a single set of attention scores for mask prediction. We observe that FPN features are beneficial for improving the segmentation performance, especially objects with varying sizes being presented. Method c adds SAM to the baseline but uses the original 36 channels of FPN features (as discussed in Section \ref{maskHeadSection}). Since SAM introduces more non-linearity in mask predictions, it obtains further improvements from Method b in accuracy, but its speed is slower as it uses more feature channels for predictions. The most improvement over the baseline comes from Model e as it includes our important components BMM and SAM as well as GIoU loss. Compared to the baseline, our full model (Model e) obtains an absolute gain of 5.2\% in accuracy while still maintains a fast inference.\\ 
\indent \textbf{Box divisions:} As discussed in Section \ref{maskHeadSection}, our SAM dynamically divides the bounding box into different sub-regions and yields a set of attention scores for each sub-region. We perform a study to evaluate the impact of
changing the maximum number of sub-regions on accuracy and efficiency. Table~\ref{tab:region} shows that we keep improving accuracy by increasing the box divisions from $2\times*$ to $6\times*$. However, the accuracy improvement tends to be marginal when it comes to $8\times*$, and the inference encounters a significant speed penalty. This is because of that more sub-regions require more base masks, which increase the network parameters. Note that the $2\times*$ version is similar to SipMask~\cite{cao2020sipmask}, but our result is superior in speed (with similar accuracy) as we use dynamic divisions instead of using a fixed $2\times2$ grid for mask prediction. In default, we use $6\times*$.
\begin{table}[!ht]
    \centering
    \begin{tabular}{c|c|c|c|c}
    \toprule
    Box division&$2\times*$&$4\times*$&\textbf{6$\times*$}&$8\times*$ \\
    \midrule
  AP& 34.0 & 35.4 & \textbf{36.3} & 36.4 \\
  Time (ms)& 33.2& 42.1&\textbf{50.6}& 88.6\\
    \bottomrule
    \end{tabular}
    \caption{Impact of the maximum number
of sub-regions (i.e., $4\times*$ means the maximum divisions of the sub-regions is 4).}
    \label{tab:region}
\end{table}\\
\indent \textbf{Base mask resolutions:} As mentioned in Section \ref{maskHeadSection}, feature upsampling is of great importance to the final mask prediction. 
Results in Table~\ref{tab:resolution} confirm this argument. Without the upsampling (last row), SG-Net produces the base mask with stride-8 features and only achieves 34.9\% in  AP. If our base mask features are upsampled to $\sfrac{1}{2}$ of the input resolution (factor = 4), the accuracy performance is improved by 1.4\% from 34.9\% to 36.3\%. Particularly, the improvement on the larger IoU threshold (AP@0.75) is large (from 38.1\% to 39.6\%), which indicates that the upsampling can greatly preserve the object details. However, we also observe that increasing base resolution introduces the speed penalty. In default, we use $\sfrac{1}{2}$ of the input resolution (factor = 4) as we obtain the best accuracy.
\begin{table}[!ht]
    \centering
    \begin{tabular}{c|c|c|c|c}
    \toprule
    Factor&AP & AP@0.5 & AP@0.75 & Time (ms)\\
    \midrule
  \textbf{4}& \textbf{36.3} & \textbf{57.1} & \textbf{39.6} &\textbf{50.6}\\
  2& 35.6 & 55.9 & 38.6 & 47.3 \\
  1& 34.9 & 55.9 & 38.1 & 40.9 \\
    \bottomrule
    \end{tabular}
    \caption{Impact of base mask resolution to the accuracy. The first column is the upsampling factor which indicates the feature map resolutions (i.e., 1 is no upsampling and 2 is $\times2$ upsampling)}
    \label{tab:resolution}
\end{table}\\
\indent \textbf{Base feature locations:} We evaluate the feature sampling locations for the base mask module. By using deeper FPN features, we can improve the performance in accuracy with an affordable penalty on speed (Table~\ref{tab:featurelocation}). 
\begin{table}[!ht]
    \centering
    
    \begin{tabular}{c|c|c|c|c}
    \toprule
    Feat. 
    locat.&AP & AP@0.5 & AP@0.75&Time (ms)\\
    \midrule
  P3& 35.0 & 56.1 & 37.0&35.7 \\
  P3-P4& 35.7 & 55.9   & 38.8 &43.2 \\
  \textbf{P3-P5}& \textbf{36.3} & \textbf{57.1} & \textbf{39.6}&\textbf{50.6} \\
    \bottomrule
    \end{tabular}
    
    \caption{Performance using different feature locations for base mask generation. $P3$, $P4$, and $P5$ are features from FPN.}
    \label{tab:featurelocation}
\end{table}
\subsection{Discussions}
Compared to the two-stage methods~\cite{feng2019dual,lin2020video,liu2019spatio,mohamed2020instancemotseg}, our SG-Net has a number of improvements as discussed below.\\ 
\indent \textbf{Feature sharing:}
Since FCOS~\cite{tian2019fcos} is flexible and only use FCNs, we can craft each task network more closely.  Particularly, we implement instance-aware connections to facilitate feature sharing across tasks (detection to segmentation and tracking). Therefore, each task head in our method can be jointly optimized in an end-to-end fashion.\\
\indent \textbf{Mask quality:}
Our mask prediction is performed on each sub-region of an instance, leading to a mask with fine granularity.
In the meta-experiments on the COCO image benchmark~\cite{lin2014microsoft}, our method outperforms recent strong methods. For instance, our margins over 
SipMask~\cite{cao2020sipmask} and CondInst~\cite{tian2020conditional} are 1.3-1.8\% and 0.8-1.3\% respectively with the same backbone settings.\\
\indent \textbf{Tracking robustness:}
Using centerness for tracking offers us two advantages. First, centerness tracking is robust to difficult cases (Fig. \ref{track}). When dealing with the highly overlapped bounding boxes, our centerness tracking can robustly handle the intractable ambiguity. Two, we can craft our tracking head seamlessly with FCOS, as centerness is collaterally produced in detection. \\
\indent \textbf{Inference efficiency:}
Our one-stage design is efficient. Our mask head avoids expensive proposal generation and RoI operations. Therefore, compared to MaskTrack R-CNN~\cite{yang2019video} whose inference time is proportional
to the number of predictions, our method holds the potential for  real-time applications and the additional inference time caused by the increased predictions is marginal.
\begin{figure}[!ht]
	\centering
	\includegraphics[width=8cm]{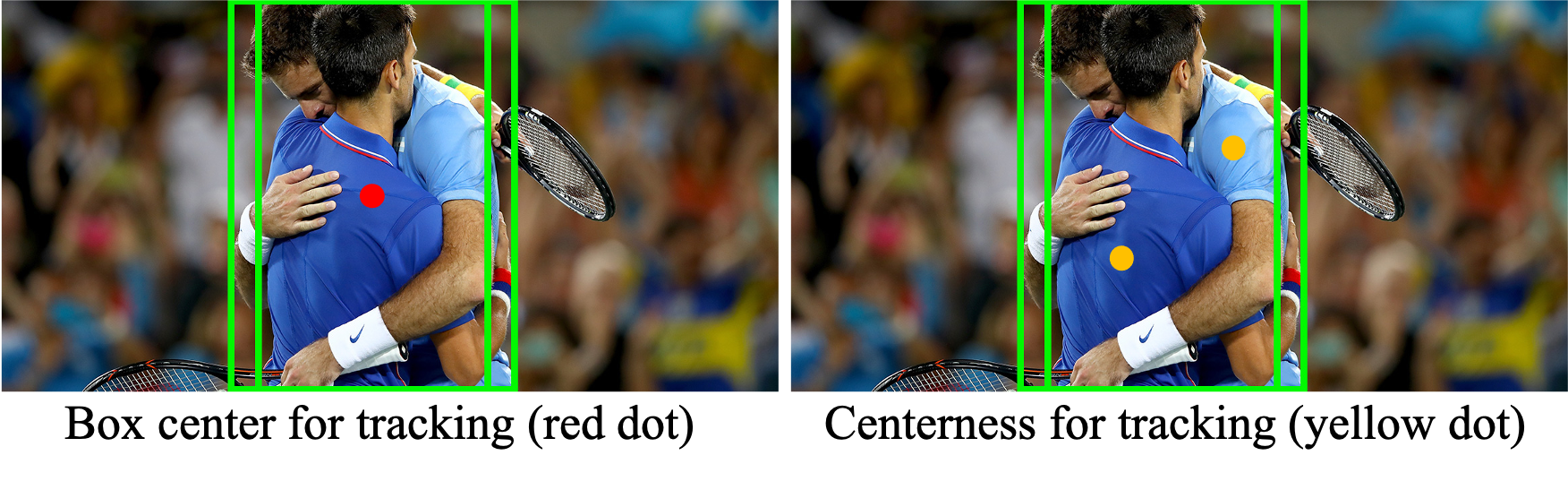}
	\caption{Box centers can reside at the same position  (left) in multiple bounding boxes, causing confusion in tracking. Our centerness tracking is robust to object appearance variation (right).}
	\label{track}
\end{figure}
\section{Conclusion}
In this study, we  propose a simple yet effective one-stage framework for VIS task, named SG-Net. Compared the two-stage solutions, we effectively improve the mask quality and inference speed. We believe that our SG-Net can be a valuable addition to the existing VIS solutions.

{\small
\bibliographystyle{ieee_fullname}
\bibliography{cvpr}
}
\end{document}